\documentclass[9pt]{article}

\usepackage{spconf}
\usepackage{placeins}
\usepackage{cite}
\usepackage{amsmath,amssymb,amsfonts}
\usepackage{algorithmic}
\usepackage{graphicx}
\usepackage{textcomp}
\usepackage{xcolor}
\usepackage{epsfig} 
\usepackage{color} 
\usepackage{lipsum}  
\usepackage{svg}
\usepackage{multirow}
\usepackage{soul}
\usepackage{subcaption}
\usepackage{hhline}
\usepackage{hyperref}

\newcommand{\fref}[1]{Fig.~\ref{#1}}
\newcommand{\sref}[1]{Section~\ref{#1}}
\newcommand{\tref}[1]{Table~\ref{#1}}
\newcommand{\eref}[1]{Eq.~(\ref{#1})}

\newcommand{\subf}[1]{#1}

\def\BibTeX{{\rm B\kern-.05em{\sc i\kern-.025em b}\kern-.08em
    T\kern-.1667em\lower.7ex\hbox{E}\kern-.125emX}}

\newcommand\blfootnote[1]{%
  \begingroup
  \renewcommand\thefootnote{}\footnote{#1}%
  \addtocounter{footnote}{-1}%
  \endgroup
}

\title{{An Optical physics inspired CNN approach for intrinsic image decomposition}}

\name{
\begin{tabular}{c}
Harshana Weligampola$^{\dagger}$ \qquad 
Gihan Jayatilaka$^{\star}$ \qquad 
Suren Sritharan$^{\dagger}$ \qquad\\
Parakrama Ekanayake$^{\star}$ \qquad 
Roshan Ragel$^{\star}$ \qquad 
Vijitha Herath$^{\star}$ \qquad
Roshan Godaliyadda$^{\star}$
\end{tabular}
}
  
\address{
    $^{\star}$ Faculty of Engineering, University of Peradeniya, Peradeniya [20400], Sri Lanka\\
    $^{\dagger}$ Faculty of IT and Computing, Sri Lanka Technological Campus, Padukka [10500], Sri Lanka
}

\begin{document}

\maketitle

\begin{abstract}
Intrinsic Image Decomposition  is an open problem of generating the constituents of an image.
Generating reflectance and shading from a single image is a challenging task specifically when there is no ground truth. 
There is a lack of unsupervised learning approaches for decomposing an image into reflectance and shading using a single image.
We propose a neural network architecture capable of this decomposition using physics-based parameters derived from the image. 
Through experimental results, we show that (a) the proposed methodology outperforms the existing deep learning-based IID techniques and (b) the derived parameters improve the efficacy significantly. We conclude with a closer analysis of the results (numerical and example images) showing several avenues for improvement.


\end{abstract}


\blfootnote{The published version of this paper:\\ H. Weligampola, G. Jayatilaka, S. Sritharan, P. Ekanayake, R. Ragel, V. Herath, R. Godaliyadda., "An Optical Physics Inspired CNN Approach for Intrinsic Image Decomposition," 2021 IEEE International Conference on Image Processing (ICIP), 2021, pp. 1864-1868, doi: \href{https://doi.org/10.1109/ICIP42928.2021.9506375}{https://doi.org/10.1109/ICIP42928.2021.9506375}.\\Correspondence: \href{mailto:harshana.w@eng.pdn.ac.lk}{harshana.w@eng.pdn.ac.lk}}
\vspace{-1mm}
\section{Introduction}
\vspace{-1mm}
Intrinsic Image Decomposition (IID) is the problem of reverting an image into its building blocks (reflectance, shading, surface normal, etc.). The reflectance depends on the material properties such as shape and color, while the shading contains information about the lighting of the environment. Information such as the geometry of the objects, shadows, directed illumination, and other ambient lights can be derived from the shading. 
Thus, extracting intrinsic features of an image is essential for various computer vision tasks. For example, using the reflectance (albedo) of an image, segmentation can be done more accurately invariant of the lighting condition \cite{lv2020attention}.
Further, tasks such as image relighting, gamma correction, and recoloring can be  accomplished using reflectance and shading information.
Therefore, it is essential to identify these intrinsic properties to guarantee the robustness of computer vision algorithms.

Many attempts have been made to decompose images into meaningful constituents. Most notably, The Retinex Theory \cite{land-retinex-theory-1977} is a biology-motivated theory based on the color constancy property of the human visual system, and it pioneered the field. Image decomposition based on this model attempts to generate reflectance and illumination maps. This has been proven useful in applications such as lighting enhancement \cite{guo2020zero,mercon-2020}. The major drawback of retinex models is their agnosticism to object surface geometries and the complicated physical phenomena related to light reflection.

The Phong illumination model \cite{phong-illumination-model} proposed a theory on imaging based on optical physics. A large body of work has been built upon this including better 3D rendering \cite{advanced-animation-and-rendering} using shading techniques. This model considers the light reflected by an object as a combination of 3 types -- ambient, specular, and diffused. Our work is based on this model.
We propose a novel Reflectance approximation map to train the neural network and a physics-based loss function to learn intrinsic properties in an image. We combine these losses  and extracted feature maps to train a neural network in an unsupervised manner. Through experimentation, we show that our model is more robust at decomposing images under a diverse set of scenes and lighting conditions compared to the existing deep learning approaches based on numerical metrics as well as sample images.

\vspace{-1mm}
\section{Related Work}
\vspace{-1mm}
IID has been attempted as a sequential algorithm, an optimization problem, and a trainable neural networks problem. Most of these approaches try to decompose in a way that reconstruction from the decomposed components is consistent with the original image. 

Classical models employ well-defined mathematical models to formulate the problem as an optimization problem, and thereby decompose the image \cite{sirfs-2017, wvm-2016, jie-2017, star-2020, bell-2014}. The main drawback of this class of models is the limitations of the mathematical definitions to capture the wide variety of imaging conditions and image capture artifacts.

Deep learning approaches try to build a model that incorporates the desirable properties of the image through loss functions. This definition is properly adapted to a wider variety of scenes by the use of large datasets \cite{lettry-2018, li2018cgintrinsics, wei2018deep, Baslamisli2020}. The limitations of these are overfitting to data, blackbox-ness, and actual decomposition diverging away from the physics.


\begin{figure}
\centering
\renewcommand{\tabcolsep}{0pt}
\renewcommand{\arraystretch}{0.0}
\begin{tabular}{c c c c c c c c}
\subf{\includegraphics[height=9mm]{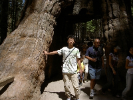}} &
\subf{\includegraphics[height=9mm]{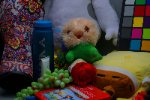}} &
\subf{\includegraphics[height=9mm]{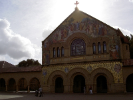}} &
\subf{\includegraphics[height=9mm]{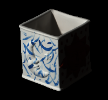}} &
\subf{\includegraphics[height=9mm]{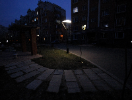}} &
\subf{\includegraphics[height=9mm]{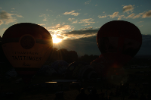}} &
\subf{\includegraphics[height=9mm]{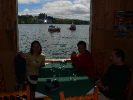}} \\

\subf{\includegraphics[height=9mm]{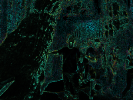}} &
\subf{\includegraphics[height=9mm]{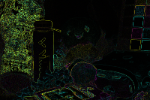}} &
\subf{\includegraphics[height=9mm]{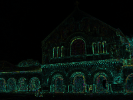}} &
\subf{\includegraphics[height=9mm]{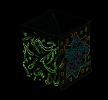}} &
\subf{\includegraphics[height=9mm]{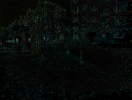}} &
\subf{\includegraphics[height=9mm]{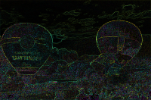}} &
\subf{\includegraphics[height=9mm]{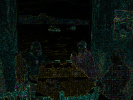}} \\

\subf{\includegraphics[height=9mm]{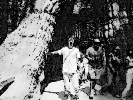}} &
\subf{\includegraphics[height=9mm]{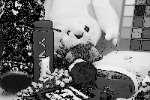}} &
\subf{\includegraphics[height=9mm]{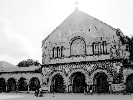}} &
\subf{\includegraphics[height=9mm]{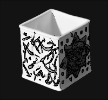}} &
\subf{\includegraphics[height=9mm]{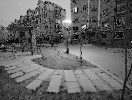}} &
\subf{\includegraphics[height=9mm]{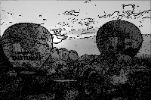}} &
\subf{\includegraphics[height=9mm]{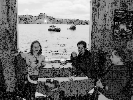}} \\

\subf{\includegraphics[height=9mm]{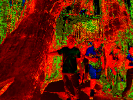}} &
\subf{\includegraphics[height=9mm]{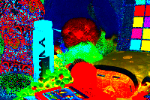}} &
\subf{\includegraphics[height=9mm]{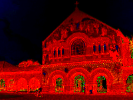}} &
\subf{\includegraphics[height=9mm]{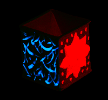}} &
\subf{\includegraphics[height=9mm]{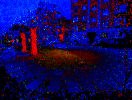}} &
\subf{\includegraphics[height=9mm]{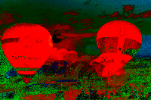}} &
\subf{\includegraphics[height=9mm]{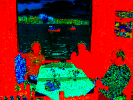}} \\

\end{tabular}

\caption{Comparison of different images (first row) with computed RRG, SG, and RAM (following rows in order)}

\label{fig:RRG_SG_RAM}
\end{figure}
\vspace{-1mm}
\section{Methodology}
\vspace{-1mm}

\subsection{Image model}

We use the Phong Reflectance model\cite{phong-illumination-model}, which is widely used to formulate image modeling. It describes a point in an image as a combination of ambient, diffuse, and specular highlights. For each light source in the scene, $i_a(\lambda)$, $i_d(\lambda)$ and $i_s(\lambda)$ are defined as the ambient, diffuse and specular intensity distribution components of the light source, where $\lambda$ is the wavelength of light. 
For multiple light sources ($\mathbf{L}$) the intensity of light reflected from a point $p$ that is represented in the image can be defined as,
\begin{equation}
    \label{eq:phong}
    \begin{split}
    I_{p} = \int_{\lambda}  k_a & r_{p}(\lambda) i_a(\lambda) + \sum_{\hat{L}^{(n)} \in \mathbf{L}} \{k_d r_{p}(\lambda) [\hat{L}^{(n)}.\hat{N}_{p}] i_{d}^{(n)}(\lambda) \\
    &+ k_s s_{p}(\lambda) [\hat{R}^{(n)}.\hat{V}]^\gamma i_{s}^{(n)}(\lambda) \} d\lambda
    \end{split}
\end{equation}
where $k_a$, $k_d$, and $k_s$ are the ambient, diffuse, and specular coefficients. 
$r_{p}(\lambda)$ is the diffuse spectral reflectance and $s_{p}(\lambda)$ is the specular spectral reflectance at point $p$. $\hat{L}_n$ is the direction vector from a point on the surface to the light direction. $\hat{N}_{p}$ is the normal at point $p$. $\hat{R}^{(n)}$ is the direction in which a perfectly reflected ray of light would travel. $\hat{V}$ is the direction pointing to the viewer. 
Note that {a hat ( $\hat{}$ ) represents that the parameter is a vector.}

We assume that the specular term is negligible in most points on the surface. Then, considering a narrow band ($\lambda_c$) we can reduce \eref{eq:phong} to, 
\begin{equation}
    \label{eq:Phong_wo_specular_int_reduced}
    I_{p}(\lambda_c) = r_{p}(\lambda_c)[k_a i_a(\lambda_c) + \sum_{\hat{L}^{(n)} \in \mathbf{L}} k_d [\hat{L}^{(n)}.\hat{N}_{p}] i_{d}^{(n)}(\lambda_c)] 
\end{equation}
Assuming that only one light source exists and that the ambient illumination is constant, we can write \eref{eq:Phong_wo_specular_int_reduced} after eliminating constant $k_d$ as,
\begin{equation}
    \label{eq:Phong_wo_specular_ambient}
    I_{p}(\lambda_c) = r_{p}(\lambda_c) [\hat{L}.\hat{N}_{p}] i_d(\lambda_c) 
\end{equation}
In \eref{eq:Phong_wo_specular_ambient}, 
we can define reflectance as $\mathbf{R} = [r_{p}(\lambda_c)]$ and shading as $\mathbf{S} = [[\hat{L}.\hat{N}_{p}] i_d(\lambda_c)]$ where $\mathbf{R}$ and $\mathbf{S}$ are matrices where each element corresponds to a pixel $p$.
Theoretically, object shape feature (normal) is included in the shading $\mathbf{S}$. But practically some of the shape features propagate to the reflectance due to the associative property of element-wise multiplication of shading and reflectance. 

\subsection{Reflectance ratio gradient (RRG)}

Consider two narrow-band channels $\lambda_{a}$ and $\lambda_{b}$. Substituting them in \eref{eq:Phong_wo_specular_ambient} we get two images in different wavelengths where each pixel is given by $I_p(\lambda_a)$ and $I_p(\lambda_b)$. As in \cite{Baslamisli2020}, we consider the natural logarithm of the ratio between $I_p(\lambda_a)$ and $I_p(\lambda_b)$.
\begin{equation}
    \label{eq:intensity_ratio}
    \mathcal{J}_p(\lambda_a,\lambda_b) = log \left(\frac{I_p(\lambda_a)}{I_p(\lambda_b)}\right) = log\left(\frac{r_{p}(\lambda_a) i_d(\lambda_a)}{r_{p}(\lambda_b) i_d(\lambda_b)}\right)
\end{equation}
For two neighboring pixels $p_1$ and $p_2$, we can assume that light intensities for a given wavelength are the same for both of these pixels (ignoring shadows). Thus, the gradient of \eref{eq:intensity_ratio} can be written as,
\begin{equation}
    \label{eq:intensity_ratio_gradient}
    \nabla \mathcal{J}(\lambda_a,\lambda_b) = \nabla log\left(\frac{r(\lambda_a)}{r(\lambda_b)}\right)
\end{equation}
    
     
    
From \eref{eq:intensity_ratio_gradient} we can show that for two given narrow wavelength bands of light, we can find the gradient of object reflectance ratio of corresponding two wavelengths.
Images have red (R), green (G), and blue (B) channels corresponding to the three wavelength bands $\lambda_R$, $\lambda_G$, and $\lambda_B$. Using \eref{eq:intensity_ratio_gradient} we get,
$\nabla \mathcal{J}(\lambda_R,\lambda_G) = \nabla log\left(\frac{r(\lambda_R)}{r(\lambda_G)}\right)$,
$\nabla \mathcal{J}(\lambda_R,\lambda_B) = \nabla log\left(\frac{r(\lambda_R)}{r(\lambda_B)}\right)$, 
$\nabla \mathcal{J}(\lambda_B,\lambda_G) = \nabla log\left(\frac{r(\lambda_B)}{r(\lambda_G)}\right)$.
These gradients are calculated for a local neighborhood using the derivative of the 2D Gaussian filter.
These three gradients are referred to as the Reflectance Ratio Gradient (RRG). 
We can use RRG to identify the boundaries of the uniform reflectance in an image.
Let $f_{RRG}$ be the function that converts an image to RRG.

\begin{figure}
    \centering
    \includegraphics[width=0.9\linewidth]{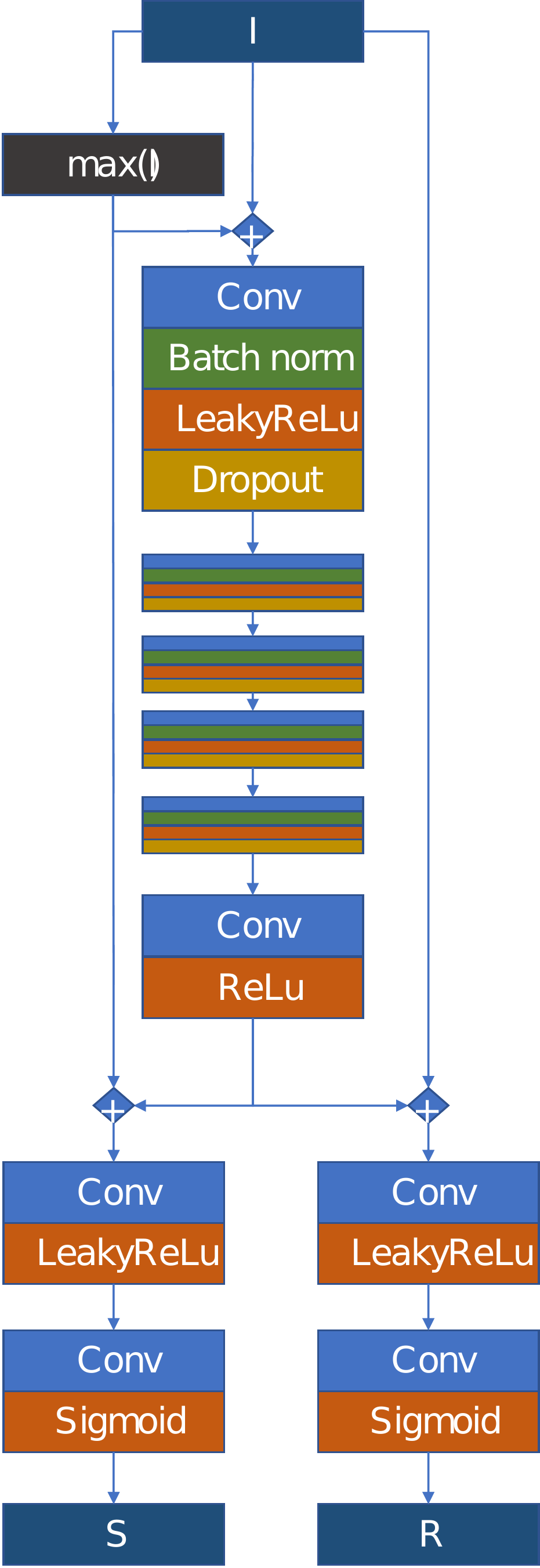}
    \caption{Illustration of the neural network architecture.}
    \label{fig:model}
\end{figure}

\subsection{Reflectance Approximation Map (RAM)}

When $r(\lambda_c)$ for all $c$ wavelengths in \eref{eq:Phong_wo_specular_ambient} are almost equal (surfaces with a shade of white), RRG will be close to zero.
Therefore, we define a reflectance approximation map that gives the likelihood of a particular channel being significant for reflectance. To define this map we first clip values from 0 to 1 in \eref{eq:intensity_ratio} which is given by $\overline{\mathcal{J}}_p(\lambda_a,\lambda_b)$. Then RAM can be given as follows, 
\begin{equation}
    \label{eq:ram}
    \resizebox{\columnwidth}{!}{
     $M_{RAM} = \left[m^{(c)}_p\right] =
    \begin{cases}
     (\overline{\mathcal{J}}_p(\lambda_R,\lambda_G) + \overline{\mathcal{J}}_p(\lambda_R,\lambda_B))/2 & \text{if }c = R\\
     (\overline{\mathcal{J}}_p(\lambda_G,\lambda_R) + \overline{\mathcal{J}}_p(\lambda_G,\lambda_B))/2 & \text{if }c = G\\
     (\overline{\mathcal{J}}_p(\lambda_B,\lambda_G) + \overline{\mathcal{J}}_p(\lambda_B,\lambda_R))/2 & \text{if }c = B
    \end{cases}
    $
    }
\end{equation}


We can use \eref{eq:ram} to identify whether the predicted reflectance is accurate. For example, if $m^{(R)}_p$ is greater than $m^{(G)}_p$ or $m^{(B)}_p$, we can say that the red channel of the albedo should be significant. Note that \eref{eq:ram} does not give actual reflectance values, but the likelihood of the reflectance. Let $f_{RAM}$ be the function that converts an image to RAM.

\subsection{Shading Gradient (SG)}

Consider a narrow band channel wavelength $\lambda_{a}$. Then take the natural logarithm of $I_p(\lambda_a)$. 
\begin{equation}
    \label{eq:intensity_log}
    \mathcal{K}_p(\lambda_a) = log \left(I_p(\lambda_a)\right)
\end{equation}
If we consider two neighboring pixels $p_1$ and $p_2$ with constant reflectance $r_{p_1}(\lambda_a) \approx r_{p_2}(\lambda_a)$, the gradient of \eref{eq:intensity_log} can be written as,
\begin{equation}
    \label{eq:intensity_log_grad}
    \begin{split}
    \nabla \mathcal{K}(\lambda_a) & = \mathcal{K}_{p_1}(\lambda_a) - \mathcal{K}_{p_2}(\lambda_a)\\
    & = \nabla log \left([\hat{L}.\hat{N}] \right)
    \end{split}
\end{equation}
As we can see from \eref{eq:intensity_log_grad}, the gradient of log intensity is equal to the gradient of normal. 

For each wavelength band, it is valid only for pixels where the reflectance corresponding to the wavelength is equal in the given neighborhood. For example, in the red channel, \eref{eq:intensity_log_grad} is valid only if $r_{p_1}(\lambda_R) \approx r_{p_2}(\lambda_R)$ where $p_1$ and $p_2$ are two pixels that are in a given neighborhood.
We can approximate that the reflectance for the red channel is equal in the given neighborhood if RRG for the red channel is low. i.e. $\nabla \mathcal{J}(\lambda_R,\lambda_G)$ and $\nabla \mathcal{J}(\lambda_R,\lambda_B)$ are low. 
Also for the green channel, if $\nabla \mathcal{J}(\lambda_G,\lambda_B)$ and $\nabla \mathcal{J}(\lambda_G,\lambda_R)$ are very low we can assume that the reflectance for the green channel is very low.
Similarly, for the blue channel, $\nabla \mathcal{J}(\lambda_B,\lambda_G)$ and $\nabla 
\mathcal{J}(\lambda_B,\lambda_R)$ should be very low.
Therefore, we can create a three-channel map using RRG where \eref{eq:intensity_log_grad} is invalid for RGB channels. This map can be given as follows.
\begin{equation}
    \label{eq:sg_invalid_map}
    \resizebox{\columnwidth}{!}{
    $
     M_{RRG} = \left[m^{(c)}_p\right] =
    \begin{cases}
     (\nabla \mathcal{J}(\lambda_R,\lambda_G) + \nabla \mathcal{J}(\lambda_R,\lambda_B))/2 & \text{if $c = R$}\\
     (\nabla \mathcal{J}(\lambda_G,\lambda_B) + \nabla \mathcal{J}(\lambda_G,\lambda_R))/2 & \text{if $c = G$}\\
     (\nabla \mathcal{J}(\lambda_B,\lambda_G) + \nabla \mathcal{J}(\lambda_B,\lambda_R))/2 & \text{if $c = B$}
    \end{cases}
    $
    }
\end{equation}
For RGB channels, we can get three gradients of normal using wavelength bands $\lambda_R$, $\lambda_G$, and $\lambda_B$ that are masked by \eref{eq:sg_invalid_map}. 
\begin{equation}
    \label{eq:intensity_log_grad_RGB}
    \nabla \mathcal{K}(\lambda_c) =
    \begin{cases}
     \nabla log \left([\hat{L}.\hat{N}] \right) & \text{if $M_{RRG}^{(c)} < 0.1$}\\
     0 & \text{otherwise}
    \end{cases}
\end{equation}
where the threshold 0.1 is selected arbitrarily.
After calculating \eref{eq:intensity_log_grad_RGB} for each RGB channel, we refer to them as the Shading Gradient (SG). Let $f_{SG}$ be the function that converts an image to SG. Visual examples of the proposed RRG,RAM and SG are illustrated in \fref{fig:RRG_SG_RAM}

\begin{figure*}[t]
\centering
\renewcommand{\tabcolsep}{1pt}
\renewcommand{\arraystretch}{-50}
\begin{tabular}{c c c c c c c c c c c c c c c}
\subf{\includegraphics[width=12mm]{fig/ORI_1.png}} &
\subf{\includegraphics[width=12mm]{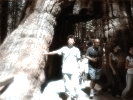}} &
\subf{\includegraphics[width=12mm]{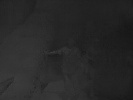}} &
\subf{\includegraphics[width=12mm]{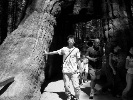}} &
\subf{\includegraphics[width=12mm]{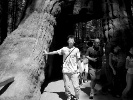}} & 
\subf{\includegraphics[width=2mm]{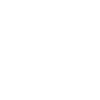}} &
\subf{\includegraphics[width=12mm]{fig/ORI_1.png}} &
\subf{\includegraphics[width=12mm]{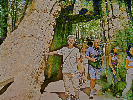}} &
\subf{\includegraphics[width=12mm]{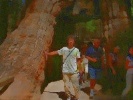}} &
\subf{\includegraphics[width=12mm]{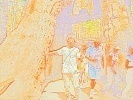}} &
\subf{\includegraphics[width=12mm]{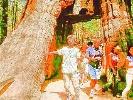}}
\\[0pt]

\subf{\includegraphics[width=12mm]{fig/ORI_3.png}} &
\subf{\includegraphics[width=12mm]{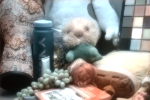}} &
\subf{\includegraphics[width=12mm]{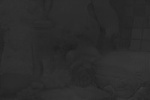}} &
\subf{\includegraphics[width=12mm]{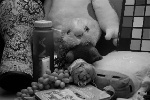}} &
\subf{\includegraphics[width=12mm]{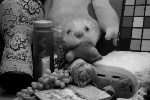}} & 
\subf{\includegraphics[width=2mm]{fig/white.jpg}} &
\subf{\includegraphics[width=12mm]{fig/ORI_3.png}} &
\subf{\includegraphics[width=12mm]{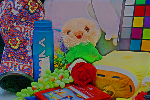}} &
\subf{\includegraphics[width=12mm]{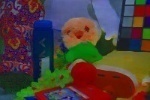}} &
\subf{\includegraphics[width=12mm]{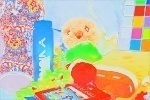}} &
\subf{\includegraphics[width=12mm]{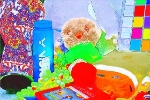}}
\\[0pt]

\subf{\includegraphics[width=12mm]{fig/ORI_4.png}} &
\subf{\includegraphics[width=12mm]{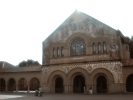}} &
\subf{\includegraphics[width=12mm]{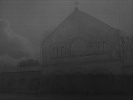}} &
\subf{\includegraphics[width=12mm]{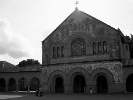}} &
\subf{\includegraphics[width=12mm]{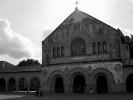}} & 
\subf{\includegraphics[width=2mm]{fig/white.jpg}} &
\subf{\includegraphics[width=12mm]{fig/ORI_4.png}} &
\subf{\includegraphics[width=12mm]{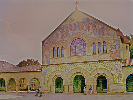}} &
\subf{\includegraphics[width=12mm]{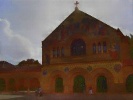}} &
\subf{\includegraphics[width=12mm]{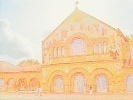}} &
\subf{\includegraphics[width=12mm]{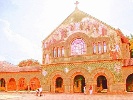}}
\\[0pt]

\subf{\includegraphics[width=12mm]{fig/ORI_6.png}} &
\subf{\includegraphics[width=12mm]{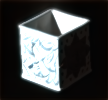}} &
\subf{\includegraphics[width=12mm]{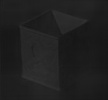}} &
\subf{\includegraphics[width=12mm]{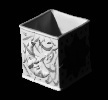}} &
\subf{\includegraphics[width=12mm]{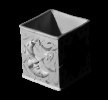}} & 
\subf{\includegraphics[width=2mm]{fig/white.jpg}} &
\subf{\includegraphics[width=12mm]{fig/ORI_6.png}} &
\subf{\includegraphics[width=12mm]{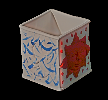}} &
\subf{\includegraphics[width=12mm]{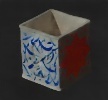}} &
\subf{\includegraphics[width=12mm]{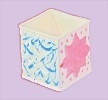}} &
\subf{\includegraphics[width=12mm]{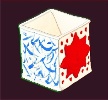}}
\\[0pt]

\subf{\includegraphics[width=12mm]{fig/ORI_7.png}} &
\subf{\includegraphics[width=12mm]{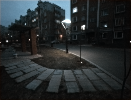}} &
\subf{\includegraphics[width=12mm]{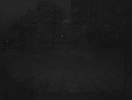}} &
\subf{\includegraphics[width=12mm]{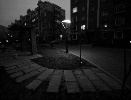}} &
\subf{\includegraphics[width=12mm]{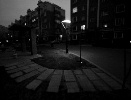}} & 
\subf{\includegraphics[width=2mm]{fig/white.jpg}} &
\subf{\includegraphics[width=12mm]{fig/ORI_7.png}} &
\subf{\includegraphics[width=12mm]{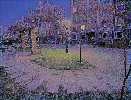}} &
\subf{\includegraphics[width=12mm]{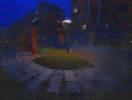}} &
\subf{\includegraphics[width=12mm]{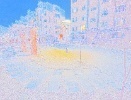}} &
\subf{\includegraphics[width=12mm]{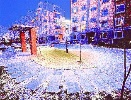}}
\\[0pt]

\subf{\includegraphics[width=12mm]{fig/ORI_8.png}} &
\subf{\includegraphics[width=12mm]{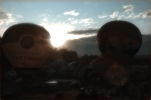}} &
\subf{\includegraphics[width=12mm]{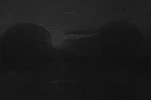}} &
\subf{\includegraphics[width=12mm]{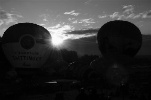}} &
\subf{\includegraphics[width=12mm]{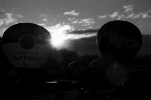}} & 
\subf{\includegraphics[width=2mm]{fig/white.jpg}} &
\subf{\includegraphics[width=12mm]{fig/ORI_8.png}} &
\subf{\includegraphics[width=12mm]{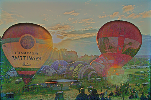}} &
\subf{\includegraphics[width=12mm]{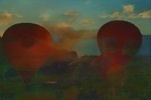}} &
\subf{\includegraphics[width=12mm]{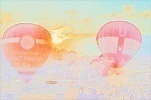}} &
\subf{\includegraphics[width=12mm]{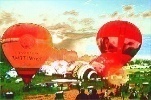}}
\\[0pt]

\subf{\includegraphics[width=12mm]{fig/ORI_9.png}} &
\subf{\includegraphics[width=12mm]{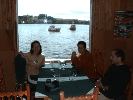}} &
\subf{\includegraphics[width=12mm]{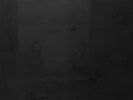}} &
\subf{\includegraphics[width=12mm]{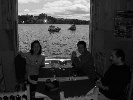}} &
\subf{\includegraphics[width=12mm]{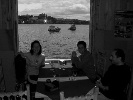}} & 
\subf{\includegraphics[width=2mm]{fig/white.jpg}} &
\subf{\includegraphics[width=12mm]{fig/ORI_9.png}} &
\subf{\includegraphics[width=12mm]{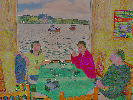}} &
\subf{\includegraphics[width=12mm]{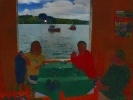}} &
\subf{\includegraphics[width=12mm]{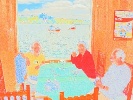}} &
\subf{\includegraphics[width=12mm]{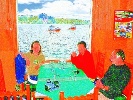}}
\\[1pt]

(a) &
(b) &
(c) &
(d) &
(e) &
&
(a) &
(b) &
(c) &
(d) &
(e) 
\\
\end{tabular}

\caption{
Comparison of shading (left) and reflectance (right) outputs: (a) Original image, (b)-(d) \cite{lettry-2018,li2018cgintrinsics,wei2018deep}, (e) Ours 
}
    
\label{fig:comparison}
\end{figure*}


\let\b\textbf
\begin{table*}
\centering
\caption{Numerical analysis on reconstructed images, reflectance (R), and shading (S). 
}
\begin{tabular}{|l|l|l|l|l|l|l|l|l|l|l|l|l|}
\hline
\multirow{2}{*}{Method} & \multicolumn{4}{l|}{LOL dataset (15 test images)}                          & \multicolumn{4}{l|}{MIT dataset (20 test images)}                    & \multicolumn{2}{l|}{MIT(R)}     & \multicolumn{2}{l|}{MIT(S)}     \\ \cline{2-13} 
                        & RMSE          & PSNR           & SSIM          & NIQE                & RMSE          & PSNR           & SSIM          & NIQE                 & RMSE           & PSNR           & RMSE            & PSNR           \\ \hline
Letry et.al             & 21.87         & \underline{35.28}    & \textbf{0.96} & 7.75                & 6.67          & \underline{39.26}    & \textbf{0.99} & {12.06}          & \textbf{41.91} & \textbf{16.58} & 40.88          & 16.46          \\ \hline
CGIntrinsic             & 63.28         & 18.95          & 0.36          & \textbf{14.78}      & 40.95         & 17.36          & 0.11          & \textbf{17.47}       & {48.47}    & \underline{16.28}    & 59.62          & 12.99          \\ \hline
Retinex-net             & \underline{6.88}    & 34.64          & 0.90          & \underline{{7.63}} & \underline{3.77}    & 37.85          & {0.95}    & \underline{{14.02}} & 67.39          & 13.48          & \underline{37.97}    & \underline{18.54}    \\ \hline
Ours                    & \textbf{2.00} & \textbf{43.12} & \underline{0.95}    & \underline{{7.63}} & \textbf{1.04} & \textbf{41.66} & \underline{0.96}    & \underline{{14.02}} & \underline{45.90}    & 15.82          & \textbf{30.54} & \textbf{20.14} \\ \hline
\end{tabular}
\label{tab:reconstruction_loss}

\end{table*}

\subsection{Intrinsic Image decomposition}
In this section, we introduce the proposed IID model, its architecture, and loss functions.

\label{sec:network_architecture}
The proposed model has one input which is the image and two outputs: a three-channel reflectance map and a single channel shading map.
The overall design of the neural network is illustrated in \fref{fig:model} and further explained in \sref{sec:experiments}. First, the input image is  concatenated with its maximum from RGB channels. 
This is connected to a sequence of five similar convolution blocks. Each convolution block consists of a reflection padding layer, a 2D Convolution layer with leaky ReLU activation function. 
The output of the final convolution block is connected to two sets of layers that will output the reflectance and shading. 
The final output layers are activated using sigmoid activation. 
The loss function used to train the model has mainly five components. They are given as follows,
\begin{equation}
    \label{eq:loss}
    \mathcal{L} = \alpha_1 \mathcal{L}_{recon} + \alpha_2 \mathcal{L}_{ss} + \alpha_3 \mathcal{L}_{rrg} + \alpha_4 \mathcal{L}_{sg} + \alpha_5 \mathcal{L}_{ram}
\end{equation}
where $\alpha_1,\alpha_2,\alpha_3,\alpha_4,\alpha_5$ are coefficients that are used to balance the loss function to train the model optimally.

\textbf{Reconstruction loss} is based on the assumption that all reflectance maps are invariable of the lighting condition. Thus, for image $i \in \mathcal{I}$ we should be able to reconstruct the original image from reflectance and shading. Therefore, reconstruction loss is given by, $\mathcal{L}_{recon} =|| \mathbf{R}_i \mathbf{S}_i - \mathbf{I}_i|| _1$




\textbf{Shading smoothness loss} ensures that the shading map is smooth where the RRG is smooth. \cite{wei2018deep} uses the reflectance map generated by the neural network itself. However, this results in color leakage to the shading map, creating a positive feedback loop which leaks texture information to the shading map.
Through RRG, we avoid such information leaks from the predicted reflectance map to the shading map, and this loss is given by, $\mathcal{L}_{ss} = || \nabla \mathbf{S}_i \exp(-10 f_{RRG}(i))|| _1$. Value 10 is chosen experimentally. These values may not be optimal.


\textbf{RRG loss} ensures the model learns the representations implied by  \eref{eq:intensity_ratio_gradient}. 
Then, RRG loss can be given as follows, $\mathcal{L}_{rrg} = || f_{RRG}(\mathbf{R}_i) - f_{RRG}(i) ||_1$


\textbf{SG loss} is used to pretrain the shape information in the shading map using \eref{eq:intensity_log_grad_RGB}. First, we reduce the channel dimension of \eref{eq:intensity_log_grad_RGB} by element-wise multiplication of each channel. Let's call the channel reduced SG function as $f'_{SG}$. Then SG loss can be given as, $\mathcal{L}_{sg} = ||(\nabla\log(\mathbf{S_i}) - f'_{SG}(i))\times f'_{SG}(i)||_1$


\textbf{RAM loss} ensures that the reflectance map is similar to the RAM except for the areas with greyscale color. Since these greyscale areas are low in RAM, we define the RAM loss as, $\mathcal{L}_{ram} = ||(\mathbf{R}_i - f_{RAM}(i))\times f_{RAM}(i)||_1$. Where $\times$ is pixel-wise multiplication.



\vspace{-2mm}
\section{Experiments}
\vspace{-1mm}
\label{sec:experiments}

This section presents experiments that were conducted to evaluate the proposed model. 
We focus on the model's ability to decompose an image into reflectance and shading.

The proposed neural network (described in Section \ref{sec:network_architecture}) contains convolution blocks with 64 filters and $3\times3$ kernels.
It is connected to a convolution layer with 32 filters which follows two parallel layers that have 16, 8, 4 filters in each convolution layer as illustrated in \fref{fig:model}.
Corresponding values for $\alpha_1,\alpha_2,\alpha_3,\alpha_4,\alpha_5$ are $1.0,0.01,0.01,0.0001,0.1$.
The model was implemented using TensorFlow\cite{tensorflow}.
We trained the neural network with the LOL dataset \cite{wei2018deep} for 100 epochs. 2000 randomly cropped image patches of size $64 \times 64$ were fed into the CNN. Each image patch was randomly augmented by flips (horizontal and vertical) and random rotations $\left(90^{\circ}, 180^{\circ}, 270^{\circ}\right)$.
The model was optimized using adam optimizer \cite{adam-optimizer} set at $\beta = 0.9$ with a learning rate of 0.002 and a decay factor of $e^{-0.01}$ for each epoch. 

The sample images in \fref{fig:comparison} were randomly selected from multiple datasets to represent a wide variety of sceneries and objects. The decomposed reflectance and shading components using our model and the prior works in \cite{lettry-2018, li2018cgintrinsics, wei2018deep} are shown in \fref{fig:comparison}.
Furthermore, the images reconstructed from these decomposed components (R \& S) were evaluated using MSE, NIQE\cite{niqe}, SSIM and, PSNR\cite{psnr-ssim} image quality assessment metrics\footnote{RMSE : lower is better. PSNR, SSIM, NIQE : higher is better.}.
These numerical metrics on LOL and MIT datasets are given in \tref{tab:reconstruction_loss} (columns 1 \& 2). In addition, the decomposed components (R, S) were also compared with the ground truth in the case of the MIT dataset (columns 3 \& 4)
\footnote{Best metric is bold and second best is underlined.}.
Through the first 2 columns, we conclude that the proposed model provides better reconstruction in comparison to other deep learning methods. Furthermore, the computed metrics on the decomposed reflectance and shading show that the proposed model consistently performs well.


\vspace{-1mm}
\section{Conclusions}
\vspace{-1mm}


In this paper, we proposed a novel image decomposition model and evaluated its performance against state-of-the-art works in image decomposition neural networks.
Through numerical evaluation metrics, we showed that the proposed model performs consistently well with different datasets consist of variety of scenes.
Though the proposed model outperforms existing works in terms of decomposition and reconstruction, there is room for improvement in regards to the color leakage problem in the shading map.


\bibliographystyle{IEEEbib}
\bibliography{main}

\end{document}